%% file: VisualSearch.tex
\documentclass[sigconf]{acmart}

\usepackage{booktabs} 
\usepackage{multirow}

\copyrightyear{2018}
\acmYear{2018}
\setcopyright{acmcopyright}
\acmConference[KDD '18]{The 24th ACM SIGKDD International Conference on Knowledge Discovery \& Data Mining}{August 19--23, 2018}{London, United Kingdom}
\acmBooktitle{KDD '18: The 24th ACM SIGKDD International Conference on Knowledge Discovery \& Data Mining, August 19--23, 2018, London, United Kingdom}
\acmPrice{15.00}
\acmDOI{10.1145/3219819.3219820}
\acmISBN{978-1-4503-5552-0/18/08}

\begin{document}
\title{Visual Search at Alibaba}




\author{Yanhao Zhang, Pan Pan, Yun Zheng, Kang Zhao, Yingya Zhang, Xiaofeng Ren, Rong Jin}
\affiliation{%
  \institution{Machine Intelligence Technology Lab, Alibaba Group}
}
\email{yanhao.zyh, panpan.pp, zhengyun.zy, zhaokang.zk, yingya.zyy, x.ren, jinrong.jr@alibaba-inc.com}




\renewcommand{\shortauthors}{Zhang et al.}

\begin{abstract}
This paper introduces the large scale visual search algorithm and system infrastructure at Alibaba. The following challenges are discussed under the E-commercial circumstance at Alibaba (a) how to handle heterogeneous image data and bridge the gap between real-shot images from user query and the online images. (b) how to deal with large scale indexing for massive updating data. (c) how to train deep models for effective feature representation without huge human annotations. (d) how to improve the user engagement by considering the quality of the content. We take advantage of large image collection of Alibaba and state-of-the-art deep learning techniques to perform visual search at scale. We present solutions and implementation details to overcome those problems and also share our learnings from building such a large scale commercial visual search engine. Specifically, model and search-based fusion approach is introduced to effectively predict categories. Also, we propose a deep CNN model for joint detection and feature learning by mining user click behavior. The binary index engine is designed to scale up indexing without compromising recall and precision. Finally, we apply all the stages into an end-to-end system architecture, which can simultaneously achieve highly efficient and scalable performance adapting to real-shot images. Extensive experiments demonstrate the advancement of each module in our system. We hope visual search at Alibaba becomes more widely incorporated into today's commercial applications.
\end{abstract}

%
%



\maketitle

\input{mainbody}

\input{experiment}

\bibliographystyle{ACM-Reference-Format}
\bibliography{visual-search-bib}
\end{document}

%% file: mainbody.tex
\vspace*{4ex}
\section{Introduction}
Visual search or content-based image retrieval (CBIR) has become a popular research topic in recent years due to the increasing prevalence of online photos in search engines and social media. Subsequently, exploiting the visual search in E-commercial systems is imperative, due to the obvious advantages 1) more convenient interaction, 2) search entry that is superior to text for fine-grained description, 3) good connection between online and offline scenarios. Considering the algorithm and engineering complexity of real-world visual search systems, there are few publications describing the end-to-end system deployed on commercial applications in detail. Generally, some of the visual search systems like Ebay~\cite{kdd_YangKBSWKP17}, Pinterest~\cite{kdd_JingLKZXDT15} release their deployed product to describe the architectures, algorithms and deployment.
\begin{figure}[t]
\begin{center}
\includegraphics[width=1.0\linewidth]{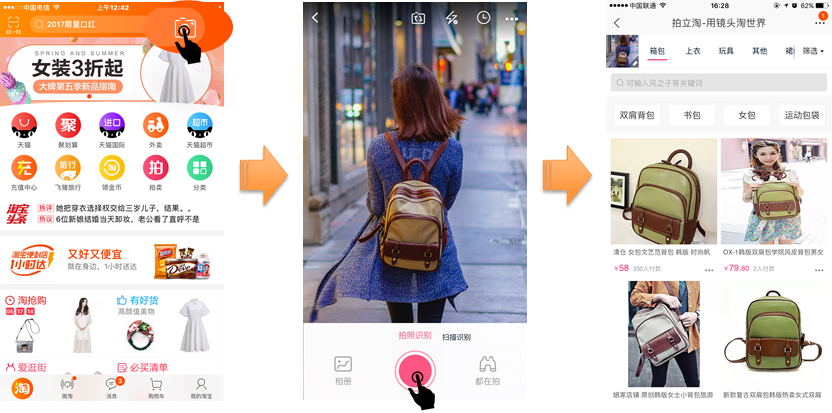}
\caption{The scenario of visual search at Alibaba\protect\footnotemark: by simply taking a picture or select any image from the photo album, ``Pailitao'' automatically returns visually similar products on Taobao marketplace and recommends even better options in real time. }
\label{fig:1_intro}
\end{center}
\end{figure}
\footnotetext{http://www.pailitao.com}

At Alibaba, we also run into many challenges when coming to practical applications of visual search technologies. By collaboration of algorithm and search teams in Alibaba, we have successfully developed an intelligence E-commercial application named ``Pailitao''. ``Pailitao'', means shopping through the camera. It is an innovative image intelligence product based on deep learning and large scale machine learning technologies as the core, and achieves the function of ``search by images'' by utilizing the visual search service, as shown in Figure~\ref{fig:1_intro}. Once launched in 2014, it triggered high attention and wide recognition in industry, and has experienced swift growth with average over 17 million Daily Active User(DAU) in 2017. On the 2017 China Double 11 Shopping festival, Pailitao successfully reached over 30 million DAU. In this paper, we would like to share some of the key developments of visual search techniques that explicitly address the existing challenges at Alibaba. These are extremely challenging and different from other products in following four major aspects:

\textbf{Heterogeneous images matching}: Unlike the standard search engines, user queries of Pailitao are usually real-shot images, which means we allow users to shot the picture from real-life or upload query images taken from any source. It is easy to notice that the quality of real-shot images is not as perfect as the inventory images, which always exist semantic and visual gaps.

\textbf{Billions of data with fine-grained categories}: Most solutions for visual search fail to operate at Alibaba scale. Alibaba has a large and continuously growing image collection, in which the labels are noisy or even wrong. In addition, the collection covers numerous fine-grained categories that are easily confused with each other. Our system needs to be both scalable and cost effective with distributed architecture to handle massive data.

\textbf{Huge expense for maintaining training data}: Noisy data always exists due to the diversity of images in a dynamic marketplace like Alibaba. 
For training deep models, these images often contain complex background and come from multiple data sources, which makes the feature learning more difficult to achieve high search relevance and low latency. Maintaining training data is laborious from collecting and cleaning data to labeling annotations, which normally requires huge cost.

\textbf{Improving the user engagement}: The success of a commercial application is measured by the benefit it brings to users. How to evolve more users to attempt the visual search service is the key issue. It is urgent to encourage them to buy the products and make possible conversions.

Despite the challenges we also have the opportunities, 1) Every item in the inventory has the own images, 2) The images are bringed with natural labeled data provided by the sellers or customers, 3) The natural scenario of shopping provides wide margin of visual search.

Seizing the existing opportunities, we describe how we alleviate the problems and address the challenges above. Overall, we present our approach in detail for building and operating visual search system at Alibaba. We illustrate the architecture of our system and take a step further to mine efficient data for feeding to deep learning model. Concretely, we describe details of how we leverage deep learning approach for category prediction and joint detection and feature learning in terms of precision and speed, along with large scale indexing and image re-ranking are discussed. We experiment on our own built test set 
to evaluate the effectiveness of each module. We also show the efficiency of our indexing engine for lossless recall and the re-ranking strategy.
\begin{figure*}[t]
\begin{center}
\includegraphics[width=0.80\linewidth]{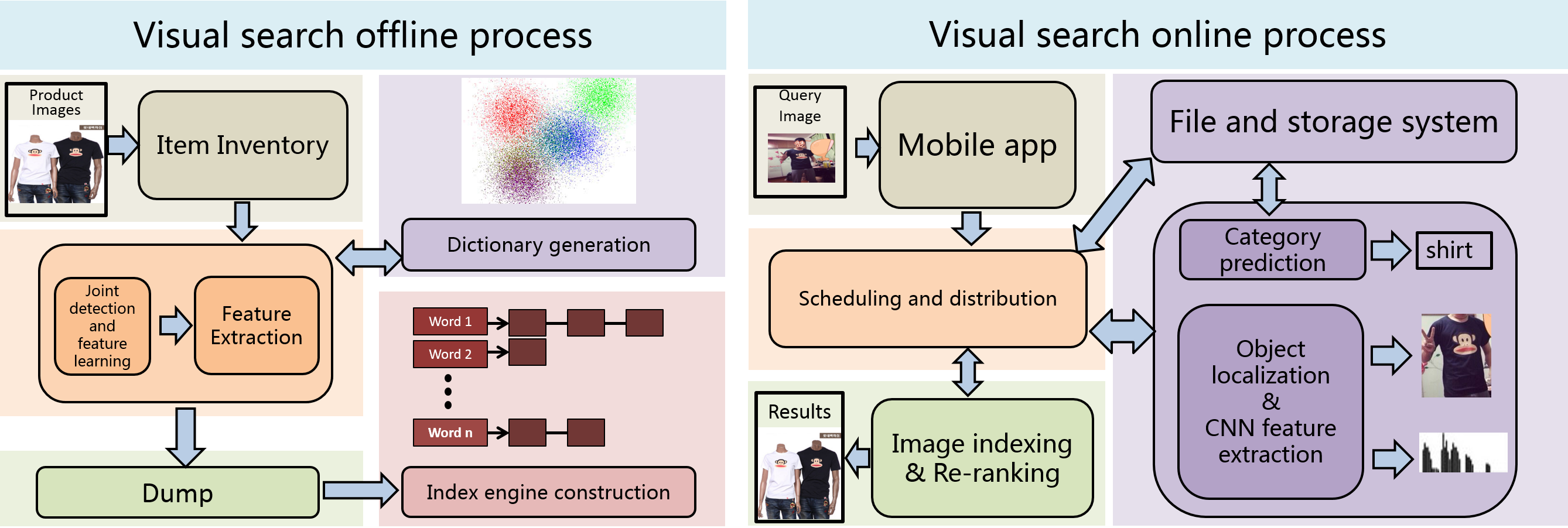}
\caption{Overview of the overall visual search architecture.}
\label{fig:6_framework}
\end{center}
\vspace*{-2ex}
\end{figure*}

\section{Related work}
Deep learning has proven extremely powerful and widely developed for semantic feature representation and image classification. With the exponential rise of deep convolutional neural networks, visual search has attracted lot of interest~\cite{kdd_BhardwajSDHPS13,kdd_JingLKZXDT15,kdd_YangKBSWKP17}. Considering the issues of large scale images that are with fine-grained categories containing complex background along with noisy labels in the practical visual search scenarios, it still remains very challenging problems on how to find the same or similar items according to query image. Taking the applied techniques into account, prior deep learning for visual search are roughly from three aspects.

\emph{CNN for instance Retrieval}: Recently, CNN~\cite{razavian2014cnn,krizhevsky2012imagenet} has exhibited promising performance for visual problems. Several works have attempted to apply CNN in image and instance retrieval~\cite{zeiler2014visualizing,Tzeng_2015_ICCV,azizpour2014factors}. By applying CNN as NeuralCode for image retrieval, e.g., Babenko et al.~\cite{eccv_BabenkoSCL14} employ the output of fully-connected layer as image feature for retrieval. Ng et al.~\cite{cvpr_NgYD15} encode CNN feature and the convolutional feature maps globally into VLAD. 
In~\cite{corr_ToliasSJ15}, Tolias et al. produces an effective visual descriptor by simply applying a spatial max-pooling over all locations on convolutional feature maps.
In our scenarios, instance retrieval differs slightly from image retrieval, because it focuses on image regions containing the target object excluding the background, rather than the entire image. 


\emph{Deep metric embedding}: Deep metric learning is proved to yield impressive performance for measuring the similarity between images. Siamese network or triplet loss is much more difficult to train in practice. To learn more effective and efficient representation, some works are designed for hard sample mining, which focuses on batch of samples that are considered hard. FaceNet~\cite{cvpr_SchroffKP15} is employed, which suggested an online strategy by associating each positive pair in the minibatch with a semi-hard negative example. 
By jointly pushing away multiple negative examples at each iteration, Sohn~\cite{nips_Sohn16} further extended the triplet loss into N-pair loss to improves triplet loss. For a massive inventory like Alibaba, the issue that the new products update frequently makes it computationally inefficient and infeasible to collect image triplets across all categories, we design the online hard sampling mining in terms of the retrieval process and user click behavior, which prove especially impressive when the images are fine-grained and various.

\emph{Weakly supervised object localization}: A number of recent works are exploring weakly supervised object localization using 
CNNs~\cite{wacv_BazzaniBAT16,pami_CinbisVS17,cvpr_OquabBLS15}. In order to localize objects, Bergamo et al~\cite{wacv_BazzaniBAT16} propose a technique for self-taught object localization involving masking out image regions to identify the regions causing the maximal activations. Cinbis et al ~\cite{pami_CinbisVS17} and Pinheiro et al ~\cite{cvpr_OquabBLS15} combine multiple-instance learning with CNN features to localize objects. 
However, these approaches yield promising results are still in multi-stages which are not trained end-to-end. Some works are required multiple network forward passes for localization, which makes it difficult to scale in practice data. Our approach is trained end-to-end to learn the object location and features of the images without strong annotations.

In spite of success of above works, there are still challenges and issues about how to settle the real product to the ground for discovering the most relevant items for user intention. Given Alibaba scale datasets, it is challenging and non-trivial to deal with billions of data and perform satisfying performance and latency.

With these realistic challenges in mind, we propose a hybrid scalable and resource efficient visual search system. We conclude our contributions as following:

1)We introduced an effective category prediction method using model and search-based fusion to reduce the search space. Compared with the traditional model-only method, our approach has better scalability and achieves better performance for confusion categories and domain restrictions.

2)We proposed a deep CNN model with branches for joint detection and feature learning. Unlike fully supervised detection methods that are trained with huge expense of human labeled data, we propose to simultaneously discover the detection mask and exact discriminative feature without background disturbance. We directly apply user click behaviors to train the model without additional annotations in a weakly supervised way.

3)As the deployed mobile application, we finish the retrieval process using binary indexing engine and re-ranking to improve the engagements. We allow users to freely take photos to find identical items with millisecond response and lossless recall in a highly available and scalable solution. Extensive experiments demonstrate the effectiveness of the end-to-end architecture of Pailitao to serve visual search for millions of users.

\begin{figure}[t]
\begin{center}
\includegraphics[width=0.80\linewidth]{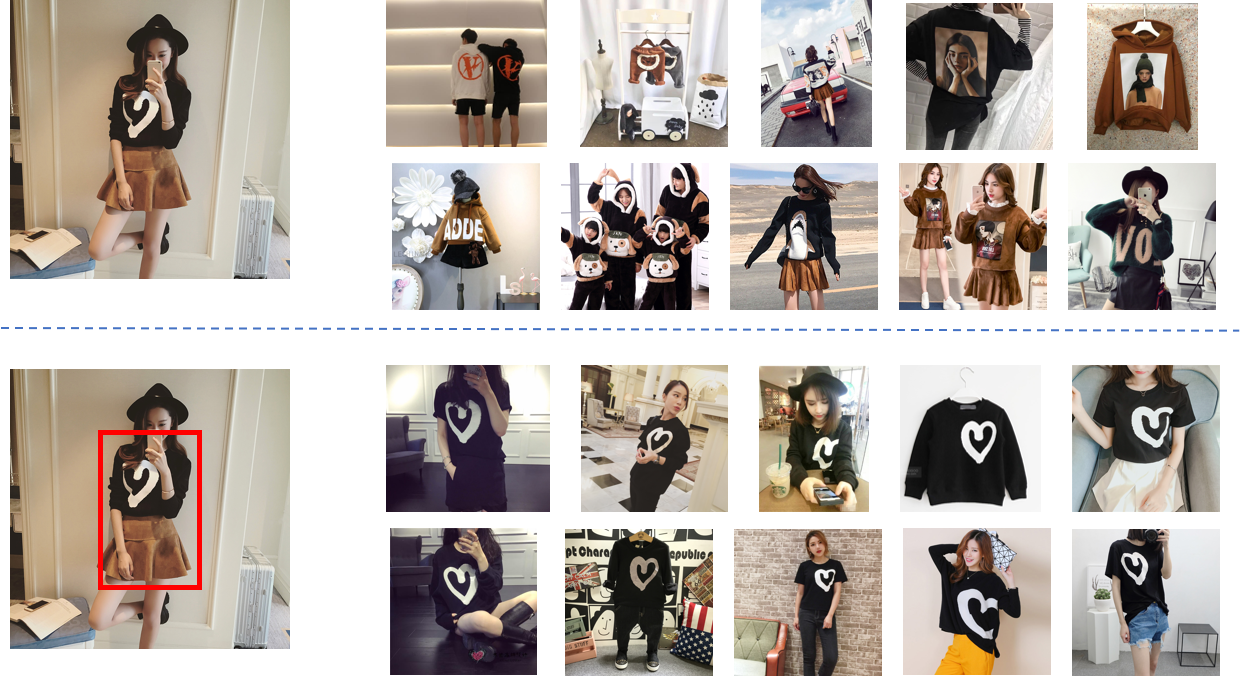}
\caption{The top row is the results without detection, which show more obvious disturbing background. Meanwhile, the bottom row shows the detected results, which have a very significant improvement and very promising.}
\label{fig:8_detcomp}
\end{center}
\vspace*{-3ex}
\end{figure}

\section{VISUAL SEARCH ARCHITECTURE}
Visual search aims at searching for images by visual features to provide users with relevant image list. As the retrieval services in terms of professional image search engine, Pailitao launched on line for the first time in 2014, by continuous polishing of product technology, it has become the application of millions of users. With the growth of business, we also settle down the stabile and scalable visual search architecture.


Figure~\ref{fig:6_framework} illustrates the overall visual search process of Pailitao, which is divided into offline and online process flow. Offline process: this mainly refers to the entire process of the building index of documents every day, involving item selection, offline feature extraction, indexing construction. After execution is completed, online inventory will be updated every day in a specified time. Online process: this mainly refers to the key steps to obtain the final result of the return process when a query image is uploaded by the user. This shares the similar process as the offline, comprising online category prediction, online detection and feature extraction. Finally, we retrieve the result list by indexing and re-ranking.

\subsection{Category Prediction}
\label{sec:catpre}
\subsubsection{Item inventory selection}
There are vast amounts of product categories and images, including the PC main images, SKU images, unboxing images and LOG images, covering all aspects of the E-commerce. We need to select the relative interesting images of the users from these massive images as item inventory to be indexed. We first filter the full gallery according to shopping preferences and image quality. For the reason that too many same or highly similar items exist on Taobao, the final search results will appear in a large number of identical items without the filter process, resulting in poor user experience. After that, we add the duplicate image remove module, which aims to remove the identical or highly similar items and optimize the indexing documents.

\begin{figure}[t]
\begin{center}
\includegraphics[width=1.0\linewidth]{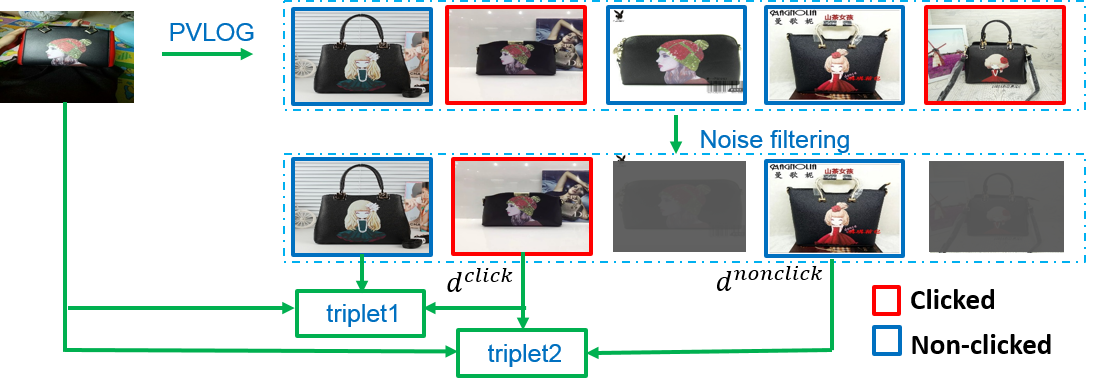}
\caption{PVLOG triplets mining strategy using user click data.}
\label{fig:13_pvlogclick}
\end{center}
\vspace*{-3ex}
\end{figure}
\subsubsection{Model and search-based fusion}
Taobao category is a hierarchy system of leaf categories, considering both certain visual and semantic similarity. Category system is not just a technical issue, but also a business problem in favor of consumer awareness. Currently, we predict 14 set categories in Pailitao to cope with user preference and narrow down the search space, covering the all leaf categories, such as shoes, dress, bags etc. For model-based part, we deploy state-of-the-art GoogLeNet V1~\cite{cvpr_SzegedyLJSRAEVR15} network for trade-off between high accuracy and low latency. The network is trained using subset of item inventory with category labels, which contains diverse product categories. As the input, each image is resized to $256\times256$ with random crop to $227\times 227$ following standard setup~\cite{cvpr_SzegedyLJSRAEVR15}. To train the network, we use the standard softmax-loss for classification task. For search-based part, we exploit the discriminative capacity of the output feature from deep network. Specifically, we collect 200 million images as the reference set with the ground truth category as pair $(x_i,y_i)$. We use binary search engine to the retrieve Top 30 results in reference set. We weight the contribution $y_i$ of each $x_i$ in 30 neighbors to predict the label $y$ of query $x$. This is based on the distances to query $x$ using the weight function $w(x,x_i)=\exp(-\lambda|x-x_i|)^2_2$, where $\lambda$ is estimated in the weight function by maximum likelihood $\lambda^*=\arg\max\Sigma^n_{i=1}log\text{Pr}(y_i|x_i;\lambda)$.

To improve the category prediction accuracy, we weighted fuse the model-based and search-based results. The validation set is also collected from the inventory data and cover all the categories.
Benefit from the distinguishing ability of the features, the search-based method correct the confused category and improve the final results. Overall, the fusion brings over 2\% absolute improvement to Top-1 accuracy in category prediction.

\subsection{Joint Detection and Feature Learning}
\label{sec:joinmodel}
In this section, we will introduce the joint detection and feature learning based on user click behavior. The main challenge under the product image search scenario is the large discrepancy between the images from consumers and sellers. The sellers' images are usually of high quality, which are shot under controlled environments with high-end camera. However, the consumers' query images are usually shot by low-end mobile phone camera and may exist uneven illumination, large blur and complex background. In order to reduce the complex background impact, it is necessary to locate the target from the image. Figure~\ref{fig:8_detcomp} reflects the user's query, demonstrating the importance of subject detection in the search results. To align the image feature between the buyers and sellers without background clutter, we propose a deep CNN model with branches based on deep metric learning to learn detection and feature representation simultaneously.

To maximum extent, we take advantage of the PV(Page View)-LOG images with the user clicked data for hard sample mining. As result, we construct valid triplets by user clicked images that is able to jointly learn object location and feature without further bounding box annotations.
\subsubsection{PVLOG triplet mining}
Specially, given an input image $q$, the first problem is to match the CNN embeddings $f(q)$ of heterogeneous images from customers and sellers reliably. It means we need to pull the distance between query image $q$ and its identical product image $q^+$ closer than the distance between query image $q$ and a different product image $q^-$. Therefore, triplet ranking loss is used as $\text{loss}(q, q^+, q^-)$:
\begin{equation}\label{eqn:triplet_loss1}
[\text{L2}(f(q),f(q^+))-\text{L2}(f(q),f(q^-))+\delta]_+
\end{equation}
where L2 denotes the normalized distance between two features and $\delta$ is the margin($\delta=0.1$). $f$ is parameterized by a CNN that can be trained end-to-end.

\begin{figure}[t]
\begin{center}
\includegraphics[width=1.0\linewidth]{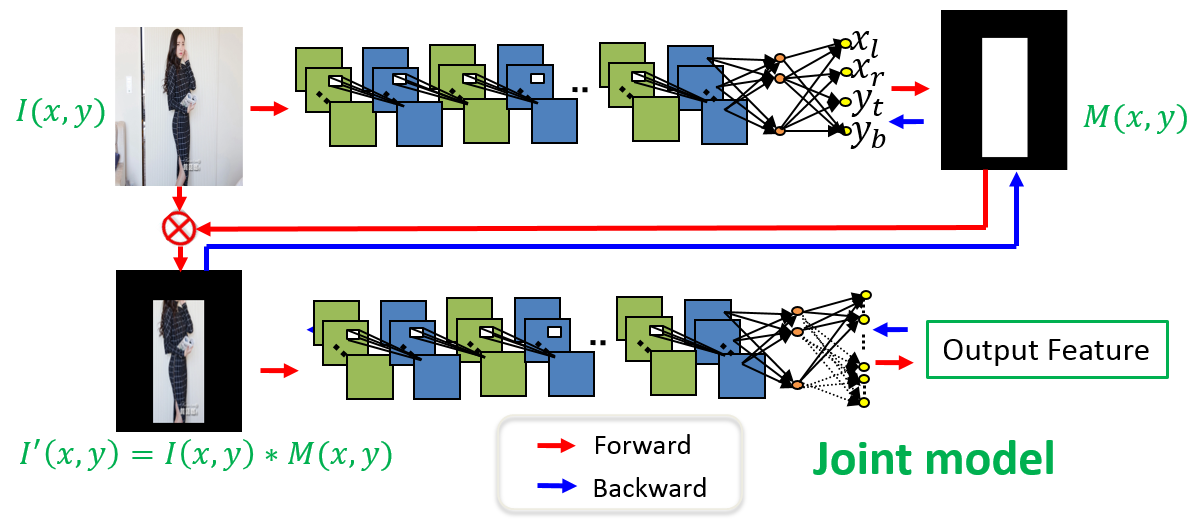}
\caption{Deep joint model with two branches for joint detection and feature learning. The top part is the detection branch. The bottom part is the feature branch.}
\label{fig:16_jointmodel}
\end{center}
\vspace*{-3ex}
\end{figure}

The main difficulty is how to obtain hard samples for training samples~\cite{cvpr_WangSLRWPCW14}. As a straightforward way, we select positive images from the same category as the query image and negative images from another category. However, positive and negative images may produce large visual difference compared with the query, which result in that triplet ranking loss easily get zero and contribute nothing during the training. Under the product image retrieval scenario in Figure~\ref{fig:13_pvlogclick}, we expect that a huge portion of the users click the identical product images from the returned list, which indicates the clicked images $d^\text{click}$ can be seen as the query's positive images. The merit of non-clicked images $d^\text{nonclick}$ is that they are usually hard negatives, meaning they are similar to query image with different product. However, non-clicked images still contain identical item to the query because the user may only click one or two of results when many identical product images return. To filter the non-clicked identical images, the negative image $q^-$ for query $q$ is computed as following,
\begin{equation}\label{eqn:1}
q^-\in \{{d^\text{nonclick}|\min[\text{dist}(d^\text{nonclick},q), \text{dist}(d^\text{nonclick},d^\text{click})] \geq \gamma}\}
\end{equation}
To compute the $\text{dist}()$ of the feature, we adopt a multi-feature fusion method by combining the local feature, previous version feature and pre-trained ImageNet~\cite{ILSVRCarxiv14} feature, which ensure noisy negatives to be found more accurately. The similar procedure is applied to click images to obtain more accurate positive images.
\begin{equation}\label{eqn:2}
q^+\in\{d^\text{click}|\text{dist}(d^\text{click},q)\leq \varepsilon\}
\end{equation}
To further expand all the available data in a mini-batch, all negative images are shared among the generated triplets in a mini-batch.
By sharing the negative samples, we can generate $m^2$ triplets before entering the loss layer compared with $m$ triplets if we don't share. To further reduce the noises in the training images, the original triplet ranking loss $\text{loss}(q, q^+, q^-)$ is improved as,
\begin{eqnarray}\label{eqn:triplet_loss}
\text{loss}&=&\frac{1}{|Q|}\sum_{q\in Q}\frac{1}{|N_q|}\sum_{q^-\in N_q}[\text{L2}(f(q),f(q^+))\\\nonumber
&&-\text{L2}(f(q),f(q^-))+\delta]_+,\\\nonumber
Q&=&\{q|\exists q^-, \text{L2}(f(q),f(q^+ ))-\text{L2}(f(q),f(q^- ))+\delta > 0\},\\\nonumber
N_q&=&\{q^-|\text{L2}(f(q),f(q^+))-\text{L2}(f(q),f(q^-))+\delta > 0\}
\end{eqnarray}
where the loss is the average computed on the query-level instead of the triplet-level, in which way we can reduce the impact of the noisy query and balance the samples. With the triplet ranking loss, we can map the buyers' real-shot images and sellers' high quality images into the same space by CNN embeddings, so that images from heterogeneous sources can be matched reliably.
\subsubsection{Unified deep ranking framework}
The second problem is to cope with the background clutter in the images. A straightforward method is deploying off-the-shelf object detection algorithms such as Faster-RCNN~\cite{pami_RenHG017} or SSD~\cite{eccv_LiuAESRFB16}. However, this approach separates the process with huge time and bounding box annotation cost that may be not optimal. We seek to jointly optimize the detection and feature learning with two branches, a deep joint model is shown in Figure~\ref{fig:16_jointmodel}.

\begin{figure}[t]
\begin{center}
\includegraphics[width=1.0\linewidth]{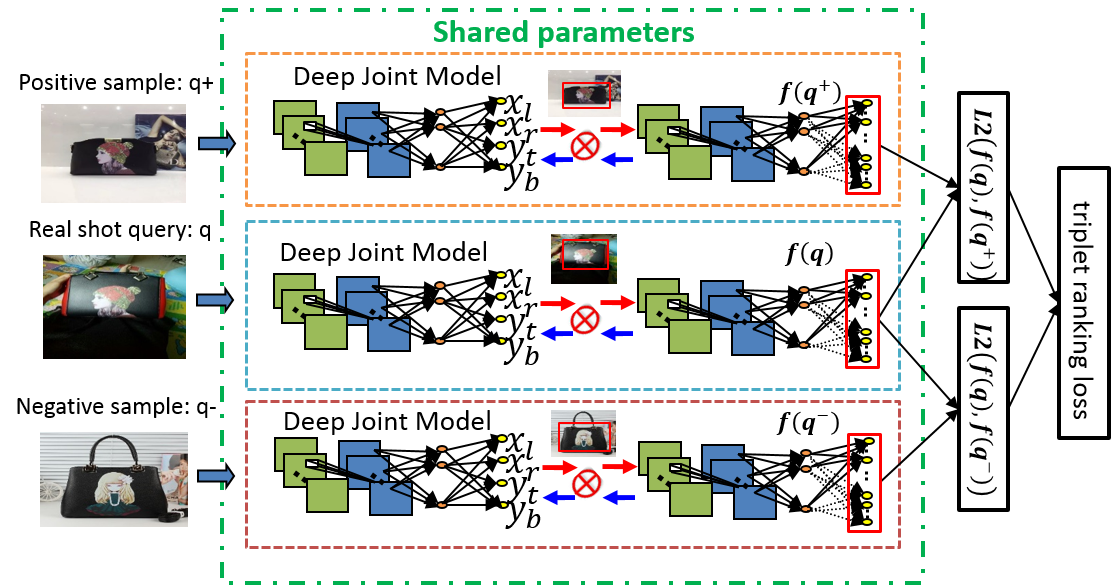}
\caption{Unified deep ranking framework consists of deep joint models for $(q, q^+, q^-)$ by feeding the triplets into the network.}
\label{fig:15_deepranking}
\end{center}
\vspace*{-1ex}
\end{figure}

We deploy deep ranking framework to learn the deep features as well as detection mask by feeding deep joint models of $(q, q^+, q^-)$ as the triplets simultaneously, maximizing the positive and negative characteristics in triplets and detecting the informative object mask without bounding box annotations. The overall deep ranking framework is shown in Figure~\ref{fig:15_deepranking}.
In each deep joint model, the detection mask $M(x,y)$ can be represented by a step function for bounding box approximation in the detection branch as shown in Figure~\ref{fig:16_jointmodel}, we element-wise multiply the image with the mask $M$ using rectangle coordinates$(x_l, x_r, y_t,y_b)$.
\begin{eqnarray}\label{eqn:triplet_loss}
&M(x,y)=[h(x-x_l)-h(x-x_r)]\times[h(y-y_t)-h(y-y_b)]\\\nonumber
&\text{where}\quad h(x-x_0)=
\left\{
\begin{array}{lr}
0, &  x < x_0  \\
1, &  x \geq x_0
\end{array}
\right.
\end{eqnarray}
However, the step function $M(x,y)$ is not differentiable. In order to perform end-to-end training, we can approximate the step function by a sigmoid function$f(x)=\frac{1}{1+e^{-kx}}$ with $k$ large enough to make it differentiable. To utilize the deep ranking framework, the triplet ranking loss addresses the region of target without background impact and encourages the discrimination of the embedding simultaneously. Notice that we only require weakly supervised user click data and do not rely on annotations of any bounding box for training, which significantly reduce the cost of human resource and improve the training efficiency.

\subsection{Image Indexing and Retrieval}

\subsubsection{Large-scale search of billion-scale images}

\begin{figure}[t]
\begin{center}
\includegraphics[width=1.0\linewidth]{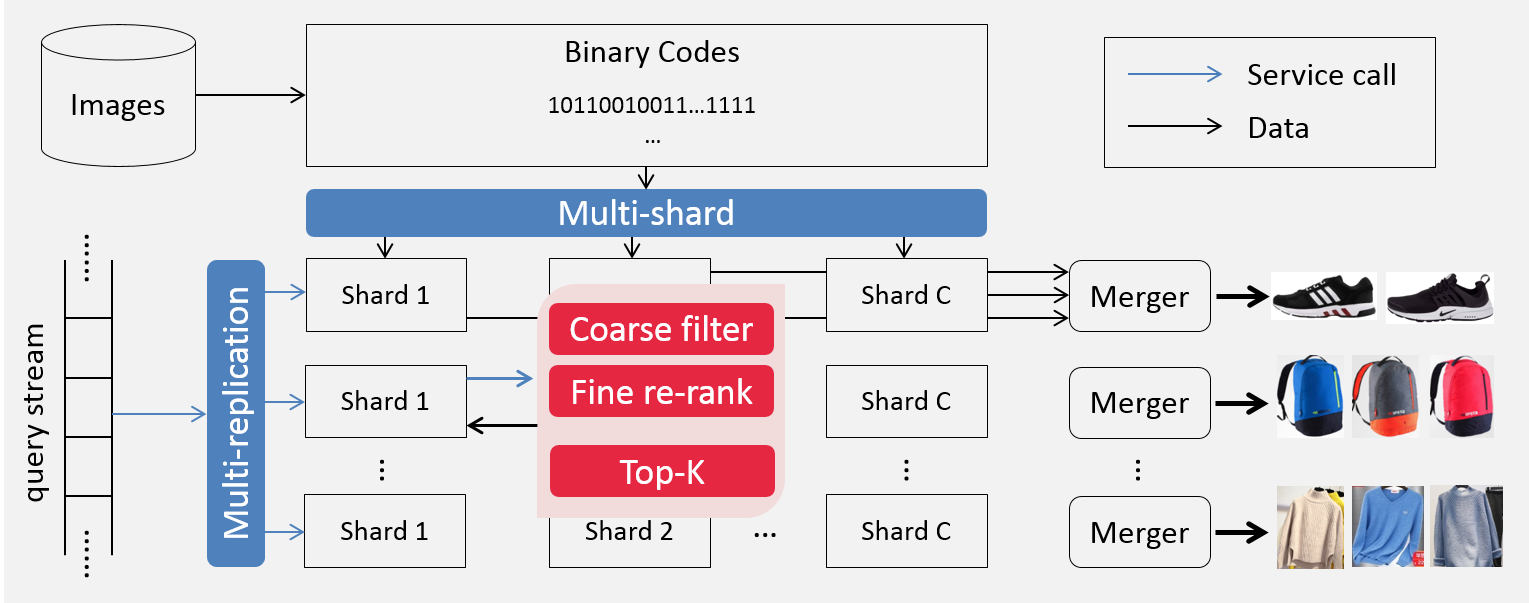}
\caption{Multi-replications and multi-shards engine architecture for image indexing.}
\label{fig:18_indexing}
\end{center}
\end{figure}

A real-time and stable search engine is very important since tens of millions of users are using the visual search service in Pailitao every day. So we adopt a multi-replications and multi-shards engine architecture as shown in Figure~\ref{fig:18_indexing}, which is not only fault-tolerant, but also very good scalability.

Multi-shards: An index instance is often difficult to store in a single machine with respect to memory and scalability. We usually use multiple machines to store the entire set of data, each shard storing only a subset of the total vectors. For a query, every shard node will search in its own subset and return its $K$ nearest neighbors. After that, a merger will be used to sort the multi-list candidates into the final $K$ nearest neighbors. Multi-shards can meet the scalability of data capacity by dynamically adding shards, and each machine only handles a fraction of vectors, helping to improve performance and recall.

Multi-replications: Query per second (qps) is an important metric for online real-time system. For Pailitao, the qps is very high, which means the latency of the search engine responses to each query is very small, posing a huge challenge to the system. Besides Alibaba has a lot of big promotions each year, it will make the qps fluctuate as much as ten times. Considering the above issues, we equip our engine with the multi-replications mechanism. Suppose there are Q queries visiting our system at the same time, we divide these queries into R parts, each part having Q/R queries. Each query part separately requests an index cluster. In this way, the number of queries that an index cluster need to process at one time decrease from Q to Q/R. With appropriate replications, we can ensure the qps not exceed the theoretical peak value.

On each node, two types of indexes are used: coarse filter and fine re-rank. The coarse filter is an improved binary inverted index constructed based on binary feature(binarization of CNN feature), with image ID as key and binary feature as value. With Hamming distance calculation, one can quickly filter out a large number of mismatched data. And then we will sort out the $K$ nearest neighbors from the returned data based on their complete binary codes. Fine re-rank is used to make a more accurate sorting refinement. It re-ranks the candidates which come from the coarse filter depending on additional metadata, such as visual attributes and local features. This process is relatively slow, partly caused by metadata stored in non-binary form, another unnegligible reason is the metadata are usually too big to locate on the memory, which means the cache hit rate is a key impact on the performance.

\subsubsection{Quality-aware image re-ranking}
We further exploit the quality-aware metadata to improve the Click Through Rate(CTR) and Click Value Rate(CVR) in order to evolve more users. Considering the initial results are obtained only by the appearance similarity, we further utilize the semantic information to re-rank the Top 60 results, including sales volume, percent conversion, applause rate, user portrait, etc. We utilize Gradient Boost Decision Tree to ensemble correlated descriptive features of different dimensions and Logistic Regression to scale the final score to $[0,1]$, which guarantee both appearance and semantic similarity and ensure that importance of each dimension can be learned. Re-ranking by quality information refines the low-quality images list with side properties while preserving the overall similarity.



%% file: experiment.tex
\section{Experiment}
In this section, we conduct extensive experiments to evaluate the performance of each module in our system. We take the GoogLeNet V1 model~\cite{cvpr_SzegedyLJSRAEVR15} as the base model for category prediction and feature learning, which follow the protocol in Section~\ref{sec:catpre} and ~\ref{sec:joinmodel}. To conduct evaluation for each component in visual search, we collected 150 thousand highest recall images along with the identical item labels of retrieved results. Our High Recall Set covers various real-shot images in 14 categories as shown in Table~\ref{tb:highrecallsetResult}. We demonstrate the end-to-end evaluation result of all components in the unified architecture with various evaluation metrics in Table~\ref{tb:highrecallsetResult}. 

\begin{table*}[t]
\caption{The end-to-end evaluation of each component on High Recall Set.}
\label{tb:highrecallsetResult}
\begin{center}
\resizebox{1.0\linewidth}{!}{%
\begin{tabular}{l|l|l||c|c|c|c|c|c|c|c|c|c|c|c|c|c||c}
\hline
Module & Component& Metric &shirt & dress & pants & bags & shoes & accessories & snacks &cosmetics& beverages & furniture & toys & underdress &digital & others & Average \\
\hline
\hline
 \multirow{3}{*}{(A)Category prediction}& model-based &Accuracy@1& 0.8163&	0.8695&	0.726&	0.9384&	0.9523	&0.9432&	0.9041	&0.9224&	0.9469&	0.9247&	0.8272&	0.83&	0.9202&	0.5952	&0.8886
\\
\cline{2-18}
&search-based &Accuracy@1&0.8651	&0.7443	&0.7644&	0.9547&	0.9666	&0.9451	&0.8365	&0.9415&	0.9249	&0.8606	&0.8225&	0.6969&	0.8282&	0.5476&	0.8551
\\
\cline{2-18}
&fusion&Accuracy@1 & 0.8042&	0.8977	&0.7781	&0.9548&	0.9809	&0.9734&	0.9104	&0.9573&	0.9615&	0.9284&	0.8781	&0.8399&	0.9387&	0.5476&	0.9101
\\
\hline
\hline
\multicolumn{2}{l|}{\multirow{3}{*}{(B)Joint detection and feature learning}}&Identical Recall@1&0.464 & 0.498&	0.393	&0.66	&0.434	&	0.224&	0.541	&0.621&	0.452	&0.267&	0.511	&0.17	&0.349&	0.439&	0.465\\
\cline{3-18}
\multicolumn{2}{l|}{}&Identical Recall@4 &	0.56	&0.616	&0.526	&0.743&	0.583&0.35&0.6	&0.716&	0.546	&0.37&	0.603	&0.2	&0.446&	0.517&	0.564\\
\cline{3-18}
\multicolumn{2}{l|}{}&Identical Recall@20 &	0.617	&0.687	&0.609	&0.781&	0.688&0.489	&0.628&	0.75	&0.6&	0.437	&0.669	&0.31&	0.532&	0.566&0.629\\
\hline
\hline
\multirow{3}{*}{(C)Indexing and retrieval} & \multirow{2}{*}{indexing engine} &Linear Recall@1 & 99.5\%
&	99.87\%&	99.88\%&	99.88\%&	100\%&	100	\%&100\%	&100\%&100\%&	99.83\%&100\%&	100\%	&100\%&97.99\%&-\\
\cline{3-18}
 &&Linear Recall@10 & 99.27\%&	99.92\%&	99.92\%&	99.71\%&	99.99\%&	99.91\%&100\%&99.97\%&100\%&99.98\%&99.99\%&	100\%&100\%&97.6\%&-\\
\cline{3-18}
  &&Linear Recall@60 & 98.68\%&	99.79\%&	99.83\%&	99.52\%&99.98\%&99.86\%&100\%&99.96\%&100\%&99.98\%&99.99\%&100\%&100\%&96.47\%&-\\
\cline{2-18}
&re-ranking & CVR & +8.45\%&+7.35\%&+4.25\%&+8.55\%&+10.15\%&	+7.54\%	&+6.49\%&+8.34\%	&+9.45\%&+10.21\%	&+7.19\%&+6.23\%	&+9.17\%	&+6.63\%	&+7.85\%
\\
\hline
\end{tabular}}
\end{center}
\vspace*{-1ex}
\end{table*}


%
%

\begin{figure}[t]
\begin{center}
\includegraphics[width=1.0\linewidth]{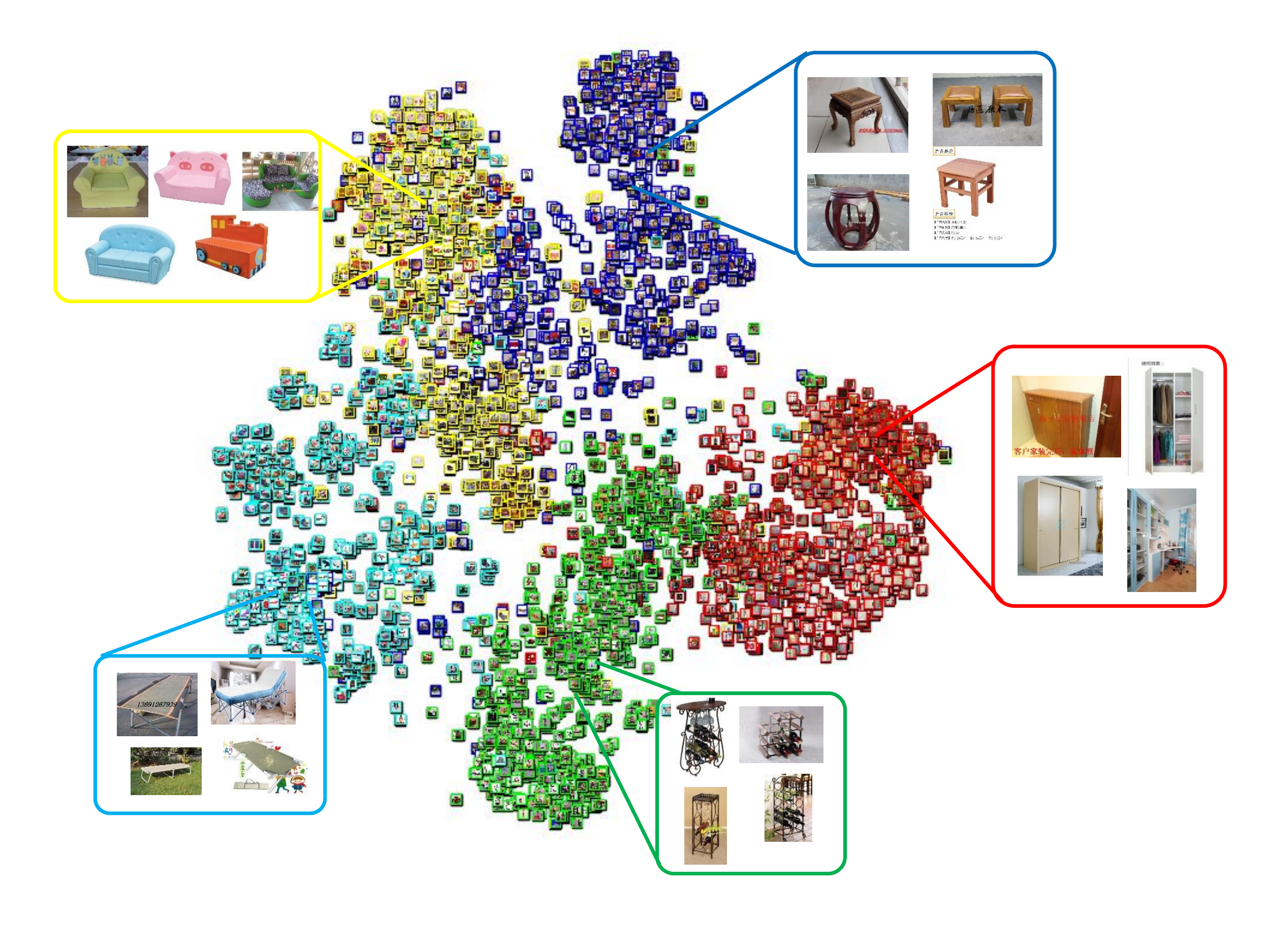}
\caption{t-SNE visualization~\cite{van2008visualizing} of 512-dim semantic feature for 5 leaf furniture categories (best viewed in color).}
\label{fig:tsne}
\end{center}
\vspace*{-1ex}
\end{figure}
\subsection{Evaluation of Category Prediction}
We conduct experiments to evaluate the performance of our fusion approach against model-based and search-based method. In Table~\ref{tb:highrecallsetResult}(A), we show that our fusion approach results in better category prediction in terms of classification Accuracy@1. Our search-based model achieves average Top-1 Accuracy 85.51\%, which is slightly lower than model-based 88.86\%. However, search-based method achieves much higher results than model-based method in some categories, i.e., shirt, pants, bags. Overall, we report the Accuracy@1 result $91.01\%$ of our fusion approach for category prediction, which increase the model-based method by $2.15\%$. The results demonstrate the complementary property of model-based and search-based methods and the fusion corrects some misclassifications of model-based method.

\subsection{Evaluation of Search Relevance}
\emph{Effect of feature branch}: To evaluate the performance of feature learning, we use the High Recall Set as the query set to search for similar images in the item inventory. We evaluate the learned feature by measuring the search relevance. Recall@K of the identical item by varying the number of top K returned results is used as Identical Recall metric. This means the query is considered to be correctly classified if there is at least one returned image belonging to the identical item among the top $K$ retrieved results. The metric measures the number of the relevant result, as the identical item will produce the most possible conversions. As the baselines, we perform start-of-the-art model-based results on the images. Our initial experiment utilized features from the original generic model (pretrained for ImageNet)~\cite{krizhevsky2012imagenet,cvpr_SzegedyLJSRAEVR15,cvpr_HeZRS16}. We computed Identical Recall@K(K=1,4,20) based on the last FC layer activations of the model. Table~\ref{tb:compareFeat} shows the Identical Recall performances of these models. Instead of deep joint model, we only select the feature branch acting on entire image that is fine-tuned by our data, which shows significant improvements compared with others.

Also, we report the overall feature results of the deep joint model on all categories in Table~\ref{tb:highrecallsetResult}(B). For all experiments, we search for 20 similar images within each predicted category for Recall@K(K=1,4,20). Identical Recall of our approach improves as $K$ increases, which clearly shows that our approach does not introduce many irrelevant images into the top search results. Compared to single feature branch, we are able to achieve better performance when retrieving with joint detection and feature learning model. The joint model suppresses the background interference and outperforms all baseline variants in all categories.
\begin{table}[t]
\caption{Comparisons of different visual features in High Recall Set.}
\label{tb:compareFeat}
\begin{center}
\resizebox{0.8\linewidth}{!}{%
\begin{tabular}{l|c|c|c}
\hline
Model& Recall@1 & Recall@4 & Recall@20 \\
\hline
\hline
Generic AlexNet~\cite{krizhevsky2012imagenet} & 0.023 &0.061 &0.122\\
\hline
Generic GoogLeNet V1~\cite{cvpr_SzegedyLJSRAEVR15}& 0.067 &0.103 &0.201 \\
\hline
Generic ResNet50~\cite{cvpr_HeZRS16} & 0.108 &0.134 &0.253\\
\hline
Generic ResNet101~\cite{cvpr_HeZRS16} & 0.128 &0.142 &0.281\\
\hline
\hline
GoogLeNet V1 feature branch(Ours) & \textbf{0.415}& \textbf{0.505}& \textbf{0.589}  \\
\hline
\end{tabular}}
\end{center}
\end{table}

\begin{figure}[t]
\begin{center}
\includegraphics[width=1.0\linewidth]{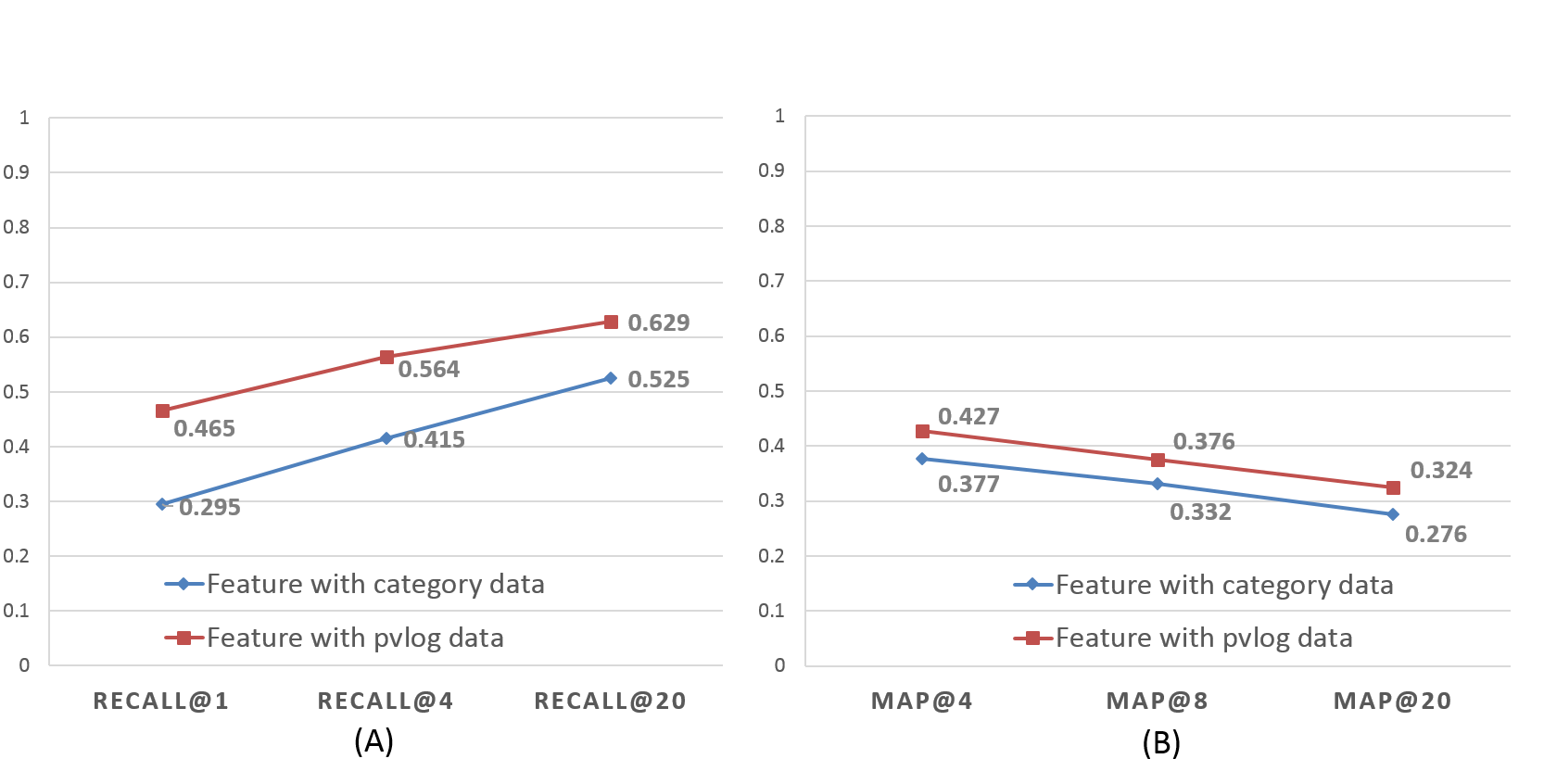}
\caption{Comparison between feature with category data and pvlog data on A) Recall (B) MAP. }
\label{fig:exp-2}
\end{center}
\end{figure}

\emph{Effect of PVLOG triplets}: As illustrated in Section~\ref{sec:joinmodel}, we found that most of clicked images are likely aimed at the identical items with query, so we train the deep features by mining the PVLOG images to form valid triplets without further annotations. To evaluate the superiority of the PVLOG triplets, we compared it with the FC layer feature of the model that are trained for category prediction using categories data. As shown in Figure~\ref{fig:exp-2}(A), we increase the Identical Recall@1 by 17 percentage point. In terms of Mean Average Precision(MAP) metric, we observe that we surpass the feature with category data by 5\% MAP@1, indicating we obtain the better and more relevant list.

\begin{figure*}[t]
\begin{center}
\includegraphics[width=1.0\linewidth]{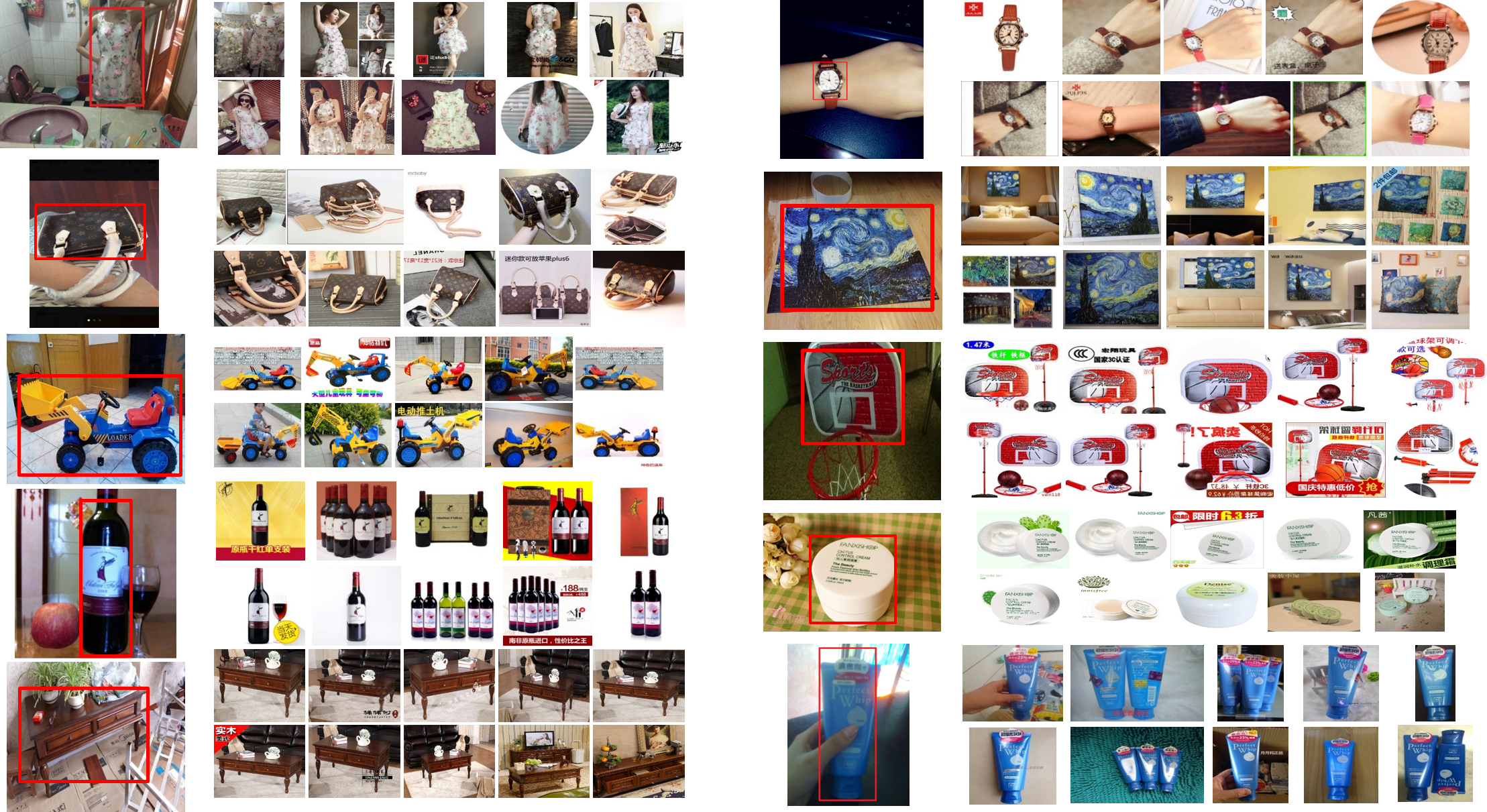}
\caption{Qualitative results of our visual search. Real-shot query images are followed by top 10 ranked images from active listings.}
\label{fig:searchRet}
\end{center}
\end{figure*}

We will further confirm the fine-grained discriminating capacity of our feature qualitatively. Figure~\ref{fig:tsne} illustrates the embeddings using tSNE~\cite{van2008visualizing} based on 512-dim semantic features of FC layer for 5 leaf furniture categories from the item inventory. This strengthens our claim qualitatively that our feature preserves semantic information and also local neighborhood. It is important to encode semantic information to mitigate the undesirable effect of collision, since the items in collision will then be semantically similar. In Figure~\ref{fig:searchRet}, we visualize the retrieval results for real-shot query images, which presents satisfying returned list on identical items.
\subsection{Evaluation of Object Localization}
\begin{figure*}[t]
\begin{center}
\includegraphics[width=1.0\linewidth]{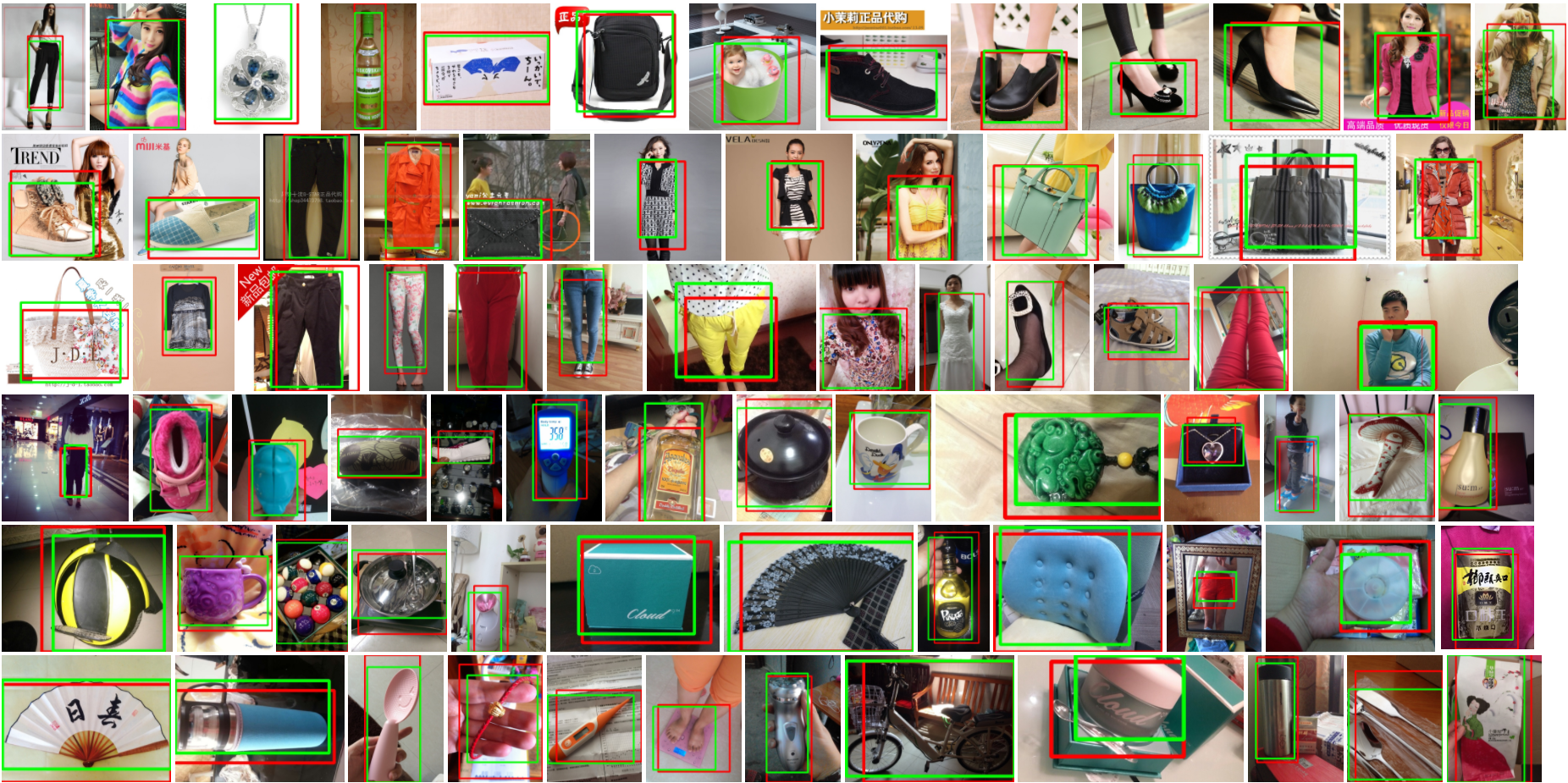}
\caption{Samples of object detection and localization results for real-shot images. Ground truth are labeled with green boxes, detected objects are in red boxes.}
\label{fig:objdet}
\end{center}
\vspace*{-1ex}
\end{figure*}

As shown in Table~\ref{tb:localizationResults}, we report that our location results of deep joint model achieves IOU@0.5 98.1\% and IOU@0.7 70.2\% compared with groundtruth bounding boxes, which are only slightly lower than fully supervised detection SSD method~\cite{eccv_LiuAESRFB16}. Figure~\ref{fig:objdet} presents the detection results of the public available images, which indicates the discriminative power of the learned detection branch and capture of object content. Meanwhile, we obtain the competitive results with much faster speed compared with the fully supervised method without compromising the Identical Recall, which achieves in a single forward pass with 20ms.

To further address the performance, we also visualize the locations of the selected objects, which localizes the fashion objects in Figure~\ref{fig:searchRet}. From these examples, we extract visual and semantic feature to get better retrieved image list. A user would easily click the identical items with query and make possible payment.

\subsection{Evaluation of Indexing and Reranking}
We show the Linear Recall for our indexing evaluation, which is utilized on 3 billion images with coarse filter of 200 thousand data in Table~\ref{tb:highrecallsetResult}(C). We compare the performance of our index result in terms of linear search, where we consider the results of the linear search as the groundtruth and evaluate how much the result approximate the groundtruth. We use Linear Recall@K to measure the quality and relevance of the rank list. The results show that we can achieve lossless recall within Linear Recall@60 compared with linear search. We also release the latency of several main components. By extensive optimization and leveraging the computational power of cloud, given a user query, it takes 30ms (model + search) on average to predict the category, and 40ms to generate image feature embedding in shopping scenarios. The ranking list takes 10ms to 20ms to return 1200 items, because coarse filter does not rely on the size of category. The quality-aware re-ranking only takes 5ms to re-rank Top 60 results. Therefore, the total latency of hundreds of millisecond, which provides users with a acceptable and enjoyable shopping experience. Furthermore, we performed the experiment that we rerank Top 60 by deploying the quality-aware re-ranking in Table~\ref{tb:highrecallsetResult}(C), achieving a relative 7.85\% increase in average CVR engagement.



\begin{table}[t]
\caption{Quantitative results of object localization compared with fully supervised detection SSD~\cite{eccv_LiuAESRFB16}.}
\label{tb:localizationResults}
\begin{center}
\resizebox{1.0\linewidth}{!}{%
\begin{tabular}{l|c|c|c|c|c|c}
\hline
Methods& IOU@0.5 & IOU@0.7 &Recall@1 & Recall@4 & Recall@20 & latency\\
\hline
\hline
Fully supervised detection SSD~\cite{eccv_LiuAESRFB16}&  98.1\% & 95.1\% &46.7\% & 56.2\% & 63.1\% & 59 ms\\
\hline
Weakly supervised detection(Ours) & 94.9\% & 70.2\% & 46.5\% &56.4\% & 62.9\% & 20 ms\\
\hline
\end{tabular}}
\end{center}
\vspace*{-1ex}
\end{table}

\section{Conclusions}
This paper introduces the end-to-end visual search system at Alibaba. We deploy effective model and search-based fusion method for category prediction. The deep CNN model with branches is designed for joint detection and feature learning by mining user click behavior without further annotations. As the mobile terminal application, we have also presented the binary index engine and discussed the way to reduce development and deployment costs and increase user engagement. Extensive experiments on High Recall Set illustrate the promising performance of our modules. Additionally, we show that our visual search solution has been deployed successfully to Pailitao, and integrated into other Alibaba internal application. In our future work, object co-segmentation and contextual constraints within images will be leveraged in Pailitao to enhance visual search relevance.